\pdfoutput=1

\documentclass[11pt]{article}

\usepackage[dvipsnames]{xcolor} 

\usepackage[final]{acl}

\usepackage{times}
\usepackage{latexsym}

\usepackage[T1]{fontenc}

\usepackage[utf8]{inputenc}

\usepackage{microtype}

\usepackage{inconsolata}

\usepackage{graphicx}

\usepackage{amsmath,amsfonts,bm}

\def\eqref#1{equation~\ref{#1}}

\def\1{\bm{1}}

\DeclareMathAlphabet{\mathsfit}{\encodingdefault}{\sfdefault}{m}{sl}
\SetMathAlphabet{\mathsfit}{bold}{\encodingdefault}{\sfdefault}{bx}{n}

\usepackage{array}
\usepackage{hyperref}
\usepackage{url}

\usepackage{booktabs}
\usepackage{cleveref}
\usepackage{makecell}
\usepackage{multirow}

\usepackage{numprint}

\newtheorem{definition}{Definition}[section]

\newcommand{\neighborhood}{\mathcal N}

\newcommand{\codensity}{\mathrm{coden}}
\newcommand{\Wasserstein}{\mathrm{W}}

\newcommand{\hiddenstate}{\mathrm{LM}}
\newcommand{\persistentimage}[1]{\mathrm{PI}^{#1}}
\newcommand{\robertabase}{\texttt{roberta-base}}

\newcommand{\RR}{\mathbb{R}}

\newcommand{\truepositive}{\color{OliveGreen}}
\newcommand{\falsepositive}{\color{BrickRed}}

\newcommand{\Faiss}{\texttt{faiss}}

\title{Local Topology Measures of Contextual Language Model Latent Spaces \\ 
With Applications to Dialogue Term Extraction}

\author{Benjamin Matthias Ruppik, 
{\bf Michael Heck}, 
{\bf Carel van Niekerk}, 
{\bf Renato Vukovic}, \\
{\bf Hsien-Chin Lin},
{\bf Shutong Feng},
{\bf Marcus Zibrowius}, 
{\bf Milica Ga\v{s}i\'{c} } \\
  Heinrich Heine University Düsseldorf, Germany \\
  \texttt{\{ruppik, heckmi, niekerk, renato.vukovic, } \\
  \texttt{linh, shutong.feng, marcus.zibrowius, gasic\}@hhu.de}
}

\begin{document}

\maketitle

\begin{abstract}
    A common approach for sequence tagging tasks based on contextual word representations is to train a machine learning classifier directly on these embedding vectors. 
This approach has two shortcomings. 
First, such methods consider single input sequences in isolation and are unable to put an individual embedding vector in relation to vectors outside the current local context of use. 
Second, the high performance of these models relies on fine-tuning the embedding model in conjunction with the classifier, which may not always be feasible due to the size or inaccessibility of the underlying feature-generation model. 
\newline
It is thus desirable, given a collection of embedding vectors of a corpus, i.e. a datastore, to find features of each vector that describe its relation to other, similar vectors in the datastore. 
With this in mind, we introduce complexity measures of the local topology of the latent space of a contextual language model with respect to a given datastore.
\newline
The effectiveness of our features is demonstrated through their application to dialogue term extraction. 
Our work continues a line of research that explores the manifold hypothesis for word embeddings, demonstrating that local structure in the space carved out by word embeddings can be exploited to infer semantic properties.
\end{abstract}

\section{Introduction}



The prevailing approach to sequence tagging tasks such as named entity recognition or dialogue term extraction involves a two-step process:
start with a general contextual vector representation for text sequences, for instance the embedding vectors created by a pretrained language model, then train a separate tagging model on top of the vector representations~\citep{lample-etal-2016-neural, ramshaw-marcus-1995-text}.
Optionally, assuming differentiability of the model and target function, one can fine-tune the representation model such that its embeddings are more suitable for the tagging task~\citep{panchendrarajan2018bidirectional}.
While highly effective, the representations may be expensive to compute, and fine-tuning a language model is not always feasible, for instance if the underlying model is hidden behind an application programming interface (API).
Thus, it is desirable to develop tagging methods which achieve the best performance on the given representations.
In fact, the performance of prompting-based approaches with large language models (LLMs) on named entity recognition tasks has lagged behind that of supervised sequence tagging approaches \citep{wang2023gptnernamedentityrecognition}.
Additionally, this leads to problems such as hallucinations and potential dataset contamination, which prevent a fair evaluation.

A more fundamental limitation of the prevailing paradigm is that the relation of a single input sequence to other sequences in the dataset cannot be taken into account.
Both the representation module and the tagging module commonly have a limited maximum context length.
They cannot process the entire dataset at once, but rather need to be provided with single sentences or paragraphs at a time.
The limited context can lead to suboptimal performance~\citep{amalvy-etal-2023-role}.
For example, consider named entity recognition for an isolated sentence such as \textit{Prince was prominently featured at the event}.
The word \textit{Prince} is ambiguous.
In a corpus containing news articles, \textit{Prince} or \textit{Prince Harry} likely appear in many articles related to the British royal family.
In a different corpus, the term \textit{Curry Prince} might appear frequently in the context of restaurant reviews.
So only with regard to the entire corpus under consideration, an informed choice on how to tag \textit{Prince} in the example sentence could be made.


In this work, we show that the relation between the representation of a single token and its containing corpus can be captured by studying the \emph{latent space} -- the collection of the language model's hidden states -- surrounding the corresponding embedding vector.
The geometry of these hidden states is known to capture both syntactic and semantic properties of the underlying text.
For instance, \citet{reif2019visualizing} find that distances between the contextual vectors of bidirectional encoder representations from transformers (BERT)~\citep{devlin-etal-2019-bert} correspond to parse tree embeddings based on the grammatical structure of the input phrases.
Here, we study neighborhoods of embedding vectors from a \emph{topological} viewpoint, and introduce descriptors of the shapes of these neighborhoods that are stable under symmetries such as  permutations, translations, and rotations.
In particular, we define descriptors based on \emph{persistent homology}, a well-established tool of topological data analysis \cite{carlsson2021topological}.

\subsection{Contribution}
\label{sec:contribution}

Consider the latent space of a language model in the neighborhood of a given contextual embedding vector.
For instance, the neighborhood of an embedding of the word \textit{cheap} in the context \textit{I am looking for options for cheap dinner} contains other occurrences of the word \textit{cheap} in different contexts, but also different words expressing a similar meaning (\textit{inexpensive}, \textit{good-value}) and words connected to the center word, such as \textit{restaurant}.
In this work, we show that:
\begin{itemize}
\item[(a)] this neighborhood contains information that is not present in the language model next-token prediction distribution, and that cannot be `distilled' into the language model via naive fine-tuning, 
\item[(b)] this additional information can be used to improve the performance of sequence tagging tasks, and
\item[(c)] this information can be efficiently summarized using low-dimensional topological feature descriptors.
\end{itemize}

Our topological descriptors are codensity at multiple scales~\cite{carlsson2008local,carlsson2014topological}, topological singularity measures based on Wasserstein norms~\cite{cohensteineredelsbrunner2010Lpstablepersistence}, and vectorized persistence modules.
Towards (a), we show that several of our one-dimensional numerical measures show minimal correlation with language model perplexity, indicating that they contain independent information.
Towards (b) and (c), we empirically demonstrate improvements on the natural language processing task of variants of term extraction. 
In each case, we build the latent space through a masked language model from a dialogue corpus.  
As a baseline, we employ a tagging model trained directly on the original language model vectors, and compare with models that take as input a combination of the language model vectors with our topological descriptors of the neighborhood within the latent space of a contextual language model.
Furthermore, we compare with models trained on features from \citet{vukovic-etal-2022-dialogue}, which are based on neighborhoods in a \emph{static} word embedding space.
We show that utilizing the \emph{contextually} augmented vectors yields statistically significant improvements.

Observation (a) is not completely new. 
For example, it is present in the idea of $k$-nearest neighbor language models \citep{Khandelwal2020Generalization, xu2023nearest}, where the current hidden state is augmented by the nearest neighbors from a datastore. 
Our low-dimensional descriptors, on the other hand, have not been deployed before, and our experiments for (b) provide the first application of contextual topological features to token level sequence tagging tasks.
Note with reference to point~(c) that summarizing a collection of vectors in a permutation-invariant way is a challenging problem in representation learning \citep{zaheer2017deepsets}, which we tackle in this work via tools from persistent homology.

Our work is complementary to other recent applications of topological methods to the study of contextual embedding spaces.
\citet{tulchinskii2023intrinsic} demonstrate that the topology of a point cloud derived from a text paragraph can be utilized in a sequence classification task, namely to differentiate human-written from artificially generated paragraphs.
Their approach takes into account solely the given paragraph's embedding vectors, and does not explore how these reside within the larger latent space.
Another approach involves constructing filtered graphs from the attention scores in a transformer model, followed by sequence-level classification based on persistent homology \citep{kushnareva-etal-2021-artificial,perez2022topological}.
However, this approach only applies to supervised sequence classification tasks, and does not yield local features required for tagging.
In a more qualitative direction, \citet{valeriani2023geometry} investigate the intrinsic dimension of the latent space through the different layers of a transformer, and
\citet{ethayarajh-2019-contextual} and \citet{cai2021isotropy} identify isolated clusters and low dimensional manifolds in the latent spaces of various language models.
However, they do not apply their quantitative local analysis to a practical task.

\section{Background and Methods}

\subsection{Latent Spaces of Contextual Language Models}


We consider the encoder part of a contextual language model, which can be thought of as a map
\[
    e \colon 
    (\RR^{d})^{\times N}
    \rightarrow
    (\RR^{h})^{\times N}.
\]
Here, $d$ is the dimension of the input layer, $h$ is the hidden dimension ($h \ll d$), and $N$ is the maximum sequence length, after which sequences will be truncated. This maximum length is usually fixed and finite.
The input of the encoder is a sequence of vectors $\mathbf{X} \in (\RR^{d})^{\times N}$ representing a tokenized context.
\emph{Tokenization} describes the process in which an input string is decomposed into a sequence of vectors.
In our setting, tokenization can be thought of as a lookup layer converting short text segments to vectors (together with position information).
Typically, longer words are split into several token vectors in this process.

The output of the encoder is a sequence of so-called hidden states. Commonly, these hidden states are inputs to the ``prediction head'' of the language model, which produces a probability distribution over the token space for the corresponding token location.

We think of a language corpus \(C\) as a collection of tokenized portions of text.
From the point of view of a language model, each instance \(i\) of a particular token appears in a specific context \(\mathbf{X}(i)\).
These contexts are filled with padding tokens so that they always have length~\(N\), permitting construction of the embedding sets \(e(\mathbf{X}(i))\).

\begin{definition}
    \label{def:ambient_corpus_embeddings_datastore}
    Given an encoder $e$ derived from a pretrained language model, the \emph{ambient corpus datastore} with respect to a corpus $C$ is the multi-set\footnote{
    We write \emph{multi-set} to allow for repetitions\slash multiplicities.
    This is relevant in our setting, because strings might appear multiple times in the corpus.}%
    \slash point cloud of all the embeddings \(e(\mathbf{X}(i))\) of all instances \(i\) of all tokens in \(C\).
\end{definition}

Note that we cannot explore the entire latent space of the language model, but only the subspace ``carved out'' by the datastore under consideration, as in \Cref{def:ambient_corpus_embeddings_datastore}.
In other words, by selecting the task-dependent ambient corpus for sampling the language model hidden states, we are making a choice of how we explore the hidden state space.
This choice of ambient corpus may have a big impact on the derived features.

\subsection{Local Topological Measures}
\label{sec:local-measures}

\begin{figure*}[t]
    \begin{center}
        \includegraphics[width=0.88\linewidth]{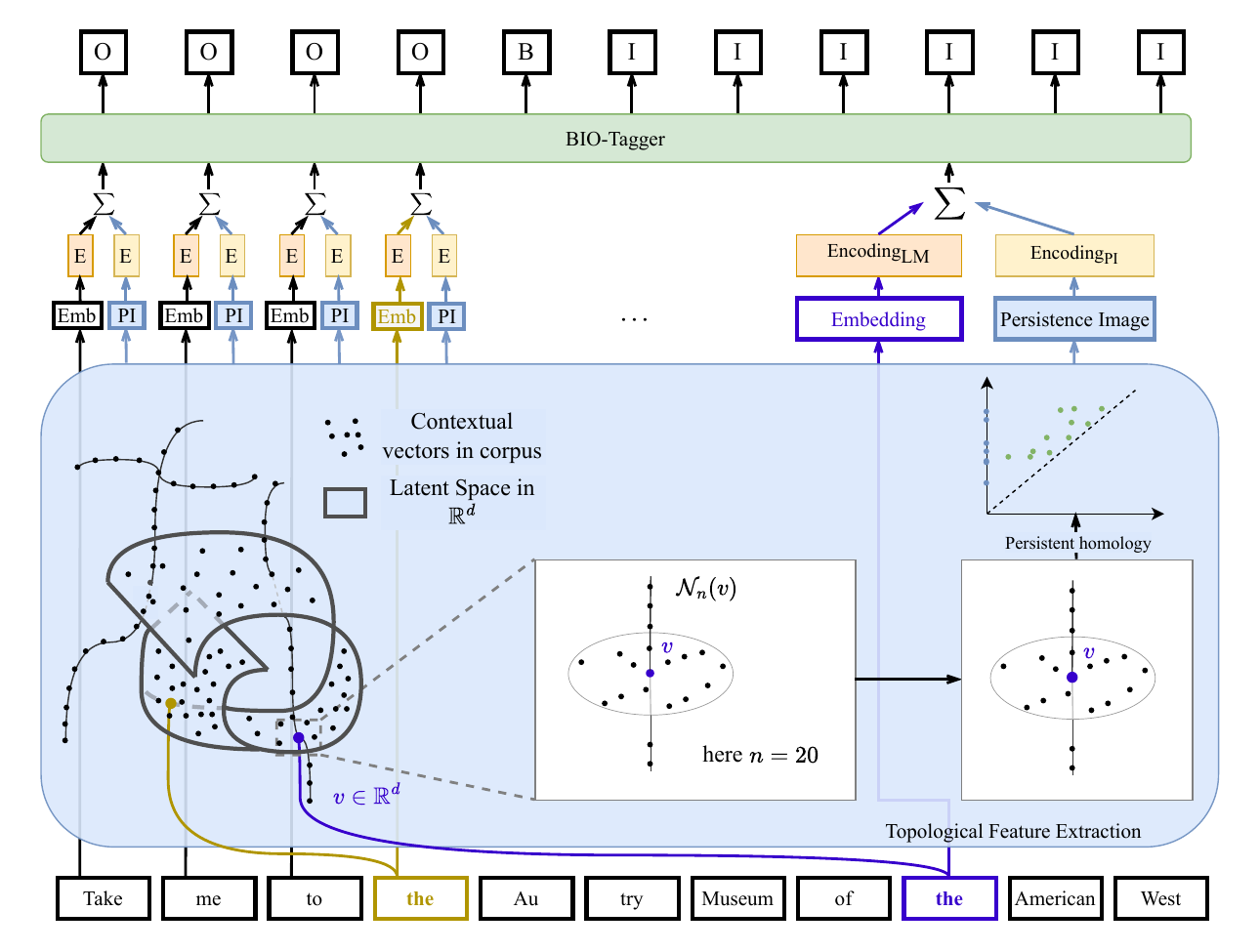}
    \end{center}
    \caption{
        \label{fig:bio_tagger_topological_deep_learning_pipeline}
        Schematic illustration of the local topological feature extraction and of our topological deep learning pipeline:
        The blue box illustrates the extraction of neighborhoods \(\neighborhood_n(v)\) in the contextualized embedding space, followed by the computation of each neighborhood's topological features, resulting in a contextualized persistence image vector.
        Note the color coding of the different occurrences of the token \texttt{'the'}; contextuality leads to different language model embedding vectors and persistence images depending on whether it is part of the term \texttt{'Autry Museum of the American West'} or used as a non-content word.
        For each token, the language model embedding (Emb) and persistence image vectors (PI) are encoded (E), combined (\(\sum\)), and then serve as input to our BIO-tagging transformer (green), which is trained on the token-level term labels (\texttt{B-TERM} (begin), \texttt{I-TERM} (inside), \texttt{O} (outside)).
    }
\end{figure*}

All our topological measures are based on neighborhoods of a given contextual embedding vector \(v\) with respect to a collection of contextual embedding vectors coming from an ambient corpus datastore.
Given an integer \(n \ge 1\), we define the \emph{neighborhood} \(\neighborhood_{n}(v)\) as the multi-set consisting of \(v\) and its $(n-1)$ nearest neighbors.
To avoid adding another copy of the query center vector \(v\) when building the neighborhood from the datastore, we first check for similarity to existing vectors in the datastore with a Euclidean distance threshold of \(10^{-4}\), and take a possible match as center vector if applicable.
For a schematic illustration of the neighborhood extraction process and the feature computation, see \Cref{fig:bio_tagger_topological_deep_learning_pipeline}.  
We consider the following local features:

\begin{description}    
    \item[Persistence Images]
    For a positive integer \(d\), \textit{persistent homology of degree \(d\)} associates with a point cloud a multi-set that encodes “\(d\)-dimensional topological features” of the cloud.
    We refer to \citet{CompTop:Intro} or \citet{Otter2017} for introductions.  
    Various vectorizations of this multi-set have been developed for subsequent use in machine learning.
    Persistence images are introduced in \citet{Adams2017} as a refined, higher-dimensional vectorization of persistent homology.
    We define \(\persistentimage{d}(v)\in\RR^{100}\) as a persistent image vector of the degree \(d\) persistent homology of \(\neighborhood_n(v)\), scaled by a factor of \(\frac{1}{n \cdot 100}\). 
    The parameter \(n\) is not included in this notation, as it will be fixed to \(128\) throughout all experiments.
    For detailed definitions, see \Cref{sec:appendix_computational_complexity_and_implementation_details}.
    The factor \(\tfrac{1}{n \cdot 100}\) appearing in our definition of the persistence image is not important at this point. 
    It is included to avoid instabilities in the training of the BIO-tagger discussed in~\Cref{sec:application}, which may otherwise arise from the vastly different scales of the coordinates of the language model embedding vectors and these additional coordinates. 
    
    \item[Wasserstein Measure]
    A simple one-dimensional vectorization is the Wasserstein norm.
    We define \(\Wasserstein^{d}_{n}(v) \in \RR\) as the Wasserstein norm of the degree~\(d\) persistent homology of \(\neighborhood_n(v)\).
    
    \item[Codensity]
    We define the $n$-th codensity \(\codensity_n(v)\in\RR\) as the 
    radius of \(\neighborhood_{n+1}(v)\).
    Higher codensity corresponds to regions where the vectors are farther spread apart.    
\end{description}

There are several reasons why we fix the cardinality \(n\) of the neighborhoods rather than, say, their radius.
Firstly, fixing the cardinality takes into account sample density of the ambient corpus from the latent space of the language model.
If we took a fixed radius, sparse regions of the ambient corpus space would be underrepresented.
Secondly, some of the topological features we consider are more readily comparable when computed on fixed cardinalities.
Indeed, a reasonable comparison of Wasserstein norms of neighborhoods of different cardinalities seems difficult, and our multiscale definition of (co)density could also not easily be emulated for neighborhoods of fixed radius.
Finally, computational complexity limits the feasibility of approaches that allow for unlimited cardinalities of neighborhoods.
For instance, in \citet{vonRohrscheidt23a}, where neighborhoods of fixed radii are employed, an additional sampling step is necessary.
More on this is discussed in \Cref{sec:appendix_computational_complexity_and_implementation_details}.

\section{Application of Local Topology Measures to Token Level Tagging Task}
\label{sec:application}

We perform a correlation analysis of local features and conduct a case study to explore the efficacy of local topology measures. 
Specifically, we apply our proposed topological feature augmentations to the task of dialogue term extraction.

\subsection{Set-Up}

\paragraph{Data}

For the term extraction case study, we resort to the MultiWOZ2.1 \citep{budzianowski-etal-2018-multiwoz,eric-etal-2020-multiwoz} and schema-guided dialogue (SGD) \citep{rastogi2020towards} task-oriented dialogue datasets.
Here, the ambient reference corpus $C$ is built solely from the language model hidden states for the training corpus of MultiWOZ2.1, so that the local measurements are comparable across both datasets.

\paragraph{BIO-Tagging}

For the sequence tagging tasks, we employ a beginning (B), inside (I), outside (O) labeling schema, as in \citet{qiu2022structure}.
To keep the comparison between our different models fair and to obtain statements about the quality of the underlying features, we choose the architectures so that the trainable BIO-tagging components have a similar number of adjustable parameters.
In this way, we can safely attribute any increase in performance to our topological augmentation of the input features rather than a stronger tagging component.

In all cases, the BIO-tagging transformer follows the RoBERTa architecture \cite{liu2019roberta} and uses 8 attention heads, 2 hidden layers, and 512 maximum position embeddings.
The language model vectors and augmenting feature vectors are fed into the BIO-tagging component through separate two-layer fully connected encoding networks with subsequent individual layer normalization, whose purpose is down- or up-scaling the feature dimension (\(768\) for the language model vectors, \(100\) for persistence images) to the hidden size (\(512\)) of the tagging transformer.
For a schematic of our BIO-tagging setup, see \Cref{fig:bio_tagger_topological_deep_learning_pipeline}.

%

\paragraph{Features}


For creating the language model embeddings, we use the second to last hidden states (at layer $11$) of the pretrained RoBERTa base model \cite{liu2019roberta}, which returns $768$-dimensional vectors, with \(L_2\)-normalization.
Note that on unit vectors, the cosine distance is proportional to the square of the Euclidean distances,
thus for the relative order in which the nearest neighbors occur, it does not matter whether we search with respect to the cosine or the Euclidean distance.

We decide on the \emph{second-to-last} hidden state of the language model, as opposed to another intermediate layer, for two reasons:
\citet{cai2021isotropy} show that the local intrinsic dimension tends to increase with the depths in the transformer, thus the resulting neighborhoods should be more expressive. 
Moreover, \citet{peters-etal-2018-deep} and \citet{tenney-etal-2019-bert} find that deeper layers in language models tend to capture more of the semantic properties, while earlier layers tend to capture the syntax.
For feature based learning, \citet{devlin-etal-2019-bert} report that among single-layer features, the second-to-last layer leads to the highest performance.
Note that our setup is not specific to the RoBERTa model or tokenizer.
Our contextualized topological features can be computed for any (masked or causal) embedding model, extraction layer, datastore produced by the model, and query dataset.

As a baseline in our term extraction experiments, we use the language model hidden state vectors described in~\Cref{sec:application} as input for the BIO-tagging model.
We test these against augmentation of the hidden states with our local persistence image descriptors introduced in~\Cref{sec:local-measures}.

\subsection{Correlation Analysis of Local Features}
\label{sec:correlation_analysis}

We begin by collecting statistical observations on the correlation between the different local topological feature types, as well as their correlation with \emph{pseudo-perplexity}, on the example of the MultiWOZ2.1 and SGD datasets.
The \emph{perplexity} of a causal language model is a model intrinsic measurement of the surprise of seeing the next token, defined as the exponentiation of the cross-entropy between the model prediction and the corpus data.
While causal perplexity is not available here, in our the masked language model setting, we apply a pseudo-likelihood score by masking tokens one by one, and computing the prediction loss of the masked out token following \citet{salazar-etal-2020-masked}.

In addition to the non fine-tuned version of the language model, here we also include the perplexity of the fine-tuned version for comparison, which uses the MultiWOZ2.1 training portion (fine-tuning for 5 epochs, \(0.15\) masking proportion, selecting the best model on MultiWOZ2.1 validation loss). 
All local features are based a non fine-tuned version of the language model.

Note that we cannot directly compute correlations between the vector-valued persistence images and the perplexity measures.
For this reason, we are relying on selected codensities \(\codensity_{n}(v)\) with values $n \in [1; 127; 511]$ and Wasserstein norms as numerical proxy estimates of the neighborhoods' topological complexity.
Our term extraction models will take only the persistence images as input, as they provide the most powerful and comprehensive representation of the local topology. 
The neural network feature extraction model can learn directly from the persistence images to estimate complexity measures approximating the Wasserstein norms, which avoids manual feature engineering.

\begin{figure}[t]
    \begin{center}
        \includegraphics[width=1.0\linewidth]{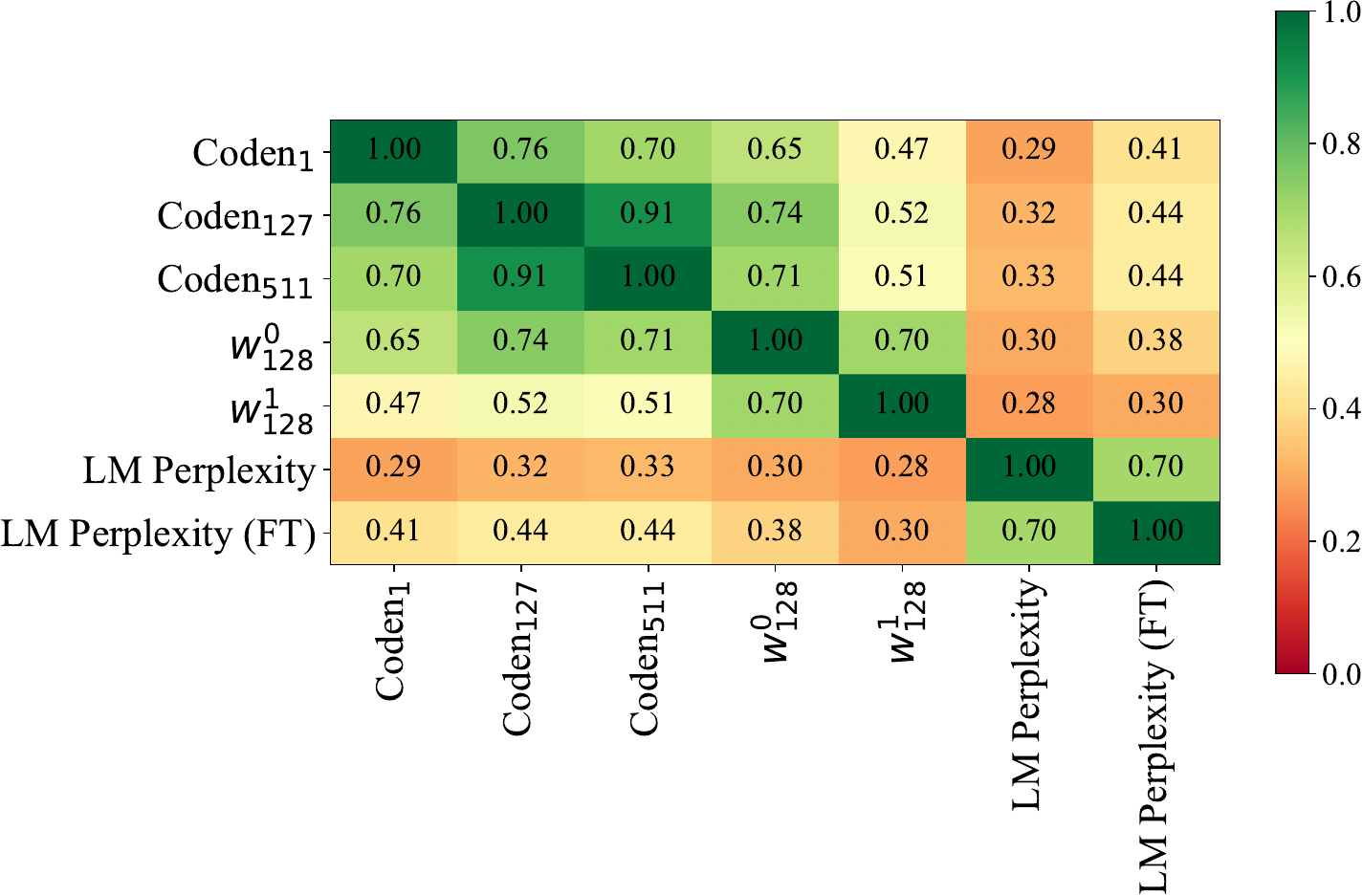}
    \end{center}
    \caption{
        Kendall's rank correlation coefficients between various local estimates and language model (LM) perplexity for the SGD test dataset.
        FT stands for LM fine-tuned on the MultiWOZ2.1 training split.
        All correlations have $p < 10^{-6}$.
        \label{fig:local_measurements_correlation_heatmap}
    }
\end{figure}

The results of Kendall's rank correlation are given in \Cref{fig:local_measurements_correlation_heatmap}.
The Wasserstein measures are strongly positively correlated with the codensities.
More importantly, the codensities and Wasserstein norms are only weakly correlated with the perplexity, indicating that these topological measures (and thus persistence images) capture information that is not present in the language model masked-token prediction distribution, and that cannot be `distilled' into the language model via naive fine-tuning.


\subsection{Case Study: Term Extraction in Task-Oriented Dialogue Data}


\subsubsection{Task Definition}

We approach dialogue term extraction as a transfer learning problem. 
Here, MultiWOZ2.1 serves as our source dataset used for training a term extractor. 
The trained model is then applied to the SGD dataset, necessitating sufficient transfer learning capabilities to properly handle the distributional shift in the data. 
We label all phrases in all utterances that match an entry in the respective dataset's ontology, i.e., that match a value in a non-categorical slot of the current turn's dialogue state or a value in the current turn's dialogue act.
The dataset ontology entities are normalized and matched to the occurrences in the respective utterances by applying the TripPy-R label map \cite{heck-etal-2022-robust} and the SGD canonical value mapping.
The ontology comprises names of entities, their domains, properties (slots), and values of these slots. 
We refer to these labelled phrases as \emph{dialogue terms}. 

These tagged spans for the dialogue datasets are encoded for the BIO-tagger, resulting in the three labels:
\texttt{O} (outside), \texttt{B-TERM}, \texttt{I-TERM} (begin and inside a term).
Since our BIO-tagging model operates on the token-level of the underlying language model, we re-align the tags with the tokenization using the IOB2 schema: 
for a word with \texttt{B}-tag, the first subtoken is tagged with \texttt{B}, its remaining subtokens with \texttt{I}.
For a word with \texttt{I}-tag, all its subtokens are tagged with \texttt{I}; and analogously for the \texttt{O}-tag.

While we employ token-level cross-entropy loss as the differentiable target function in the model training, the objective of term extraction within the context of this work is to retrieve each unique target dialogue term at least once.
That is, we do not require the tagger to find all occurrences:
We normalize the predicted phrases and ground truth by lower-casing, and deduplicate the resulting collections.
A term is considered a true positive if it is identical to exactly one ground truth term.
If a term cannot be assigned to any ground truth term, e.g., comprises several ground truth terms or is an incomplete substring of a ground truth term, it is considered a false positive.
The left-over ground truth terms without a matched prediction are the false negatives.
We call the resulting prediction, recall and F1-scores the \emph{phrasal results}.

We train the BIO-tagger for 10 epochs, using the AdamW optimizer~\citep{DBLP:conf/iclr/LoshchilovH19} with learning rate \(5 \cdot 10^{-5}\), linear warm-up for \(10\%\) of training steps and batch size of 48.
The model predictions on the held-out MultiWOZ2.1 validation set are evaluated every \(\numprint{100}\) batches for the first \(\numprint{3000}\) global steps, and the model checkpoint with the best phrasal results on the validation set is selected as the final model.

Our goal is to show that injecting our local topological features into the model yields statistically significant improvements over the original language model embeddings.
We run statistical tests over multiple different random seeds for initialization and check for significant changes in evaluation scores on the transfer set, which we take as the full collection of \numprint{463284} utterances from the SGD dataset comprising 20 domains.

\paragraph{Training Data}

In the full data setting, we train on all \(\numprint{113556}\) utterances of the MultiWOZ2.1 training split, the results are included in \Cref{tab:term_extraction_phrase_level_baseline_lm_hidden_vs_others_full_data}.

To demonstrate that the contextual topological features are useful in settings with reduced data and might help in mitigating overfitting, we create a variation of the transfer task by restricting training to subsets of the MultiWOZ2.1 dataset.
This is a more realistic transfer setting, since a good model checkpoint needs to demonstrate that it can generalize to the unseen left-out domain which it encounters in the MultiWOZ2.1 validation split for the first time.
Given one of the five major domains 
\(
    \mathcal{D} \in \left[ \text{attraction}; \text{hotel}; \text{restaurant}; \text{taxi}; \text{train} \right]
\)
in the training split, we exclude those utterances contained in dialogues from \(\mathcal{D}\) in the tagger training,
which leaves \(
    \left[ \numprint{71768} ; \numprint{59222} ; \numprint{58156} ; \numprint{86568} ; \numprint{66736} \right]
\)
utterances respectively.
Model selection is performed based on phrasal F1-score on the \numprint{14748} validation utterances, which span over all five domains.
We report results of this cross-validation setup by macro averaging the phrasal scores on the SGD dataset over 10 seeds for each of the five left out data folds in \Cref{tab:term_extraction_phrase_level_baseline_lm_hidden_vs_others_cross_validation}.

\paragraph{Static Topological Features Baseline}

We evaluate term extraction performance on the level of phrase predictions.
The phrase-level evaluation allows a comparison with~\citet{vukovic-etal-2022-dialogue}, who present a method that employs \emph{static} topological descriptors in sequence tagging tasks.
The main differences to our approach with contextual topological features are as follows:
\begin{itemize}
    \item Our local persistent homology descriptors are defined on token level with respect to the tokenization of the language model.
    This is essential for combining our new features with the language model embeddings to create fusion models which can provide predictions on token-level.
    The \emph{static} topological features of \citet{vukovic-etal-2022-dialogue} operate on word level, and they only gained contextuality in the BIO-tagging component of the model.
    Note that in this and our work, the context of an input sequence is a single dialogue utterance.
    \item \citet{vukovic-etal-2022-dialogue}'s features were based on neighborhoods in an ambient static word space composed of the \numprint{100000} most common words in the English language.
    Thus, their method depends both on having a word-level separation of the input data, and on a given dictionary.
    Our contextualized features on the other hand can be defined without any additional external data.
\end{itemize}

For a comparison between static and contextual features, we align the static topological features with the \robertabase \, tokenization in our BIO-tagging setup, and train BIO-taggers with the same architecture and data as in the contextual topological feature setting.
To that end, the first constituent subtoken of each word is augmented with the word's corresponding 100-dimensional \(H_{0}\)-persistence image feature vector of \citet{vukovic-etal-2022-dialogue}.

\subsubsection{Results}

\paragraph{Quantitative Analysis}

\begin{table}[t]
    \begin{center}
        \footnotesize
        \renewcommand{\arraystretch}{1.2}
\begin{tabular}{@{}llll@{}}
    \toprule
      \multicolumn{1}{c}{Input features} 
      & \multicolumn{1}{c}{Precision \(\uparrow\)}
      & \multicolumn{1}{c}{Recall \(\uparrow\)}
      & \multicolumn{1}{c}{F1 \(\uparrow\)}
    \\ 
    \midrule 
    \makecell{\(\hiddenstate\) \robertabase}
       & {\( 48.89 \)} 
       & {\( 56.61 \)} 
       & {\( 52.39 \)} 
    \\[0.3em] 
    \makecell{\(\hiddenstate \; \robertabase\) \\ \(\oplus \text{static } \persistentimage{0}\)}
       & {\( 49.33 \)} 
       & {\( \textbf{58.82} \)} 
       & {\( 53.62 \)} 
    \\[0.7em] 
    \makecell{\(\hiddenstate \; \robertabase\) \\ \(\oplus \text{contextual } \persistentimage{0} \)}
       & {\( \textbf{50.26}^{\star} \)} 
       & {\( 58.44 \)} 
       & {\( \textbf{53.97}^{\star} \)} 
    \\ \bottomrule 
\end{tabular}
    \end{center}
    \caption{
        \label{tab:term_extraction_phrase_level_baseline_lm_hidden_vs_others_full_data}
        Phrasal-level performance comparison for term extractors trained on the MultiWOZ2.1 training split and evaluated on the full SGD dataset. 
        Results are averages over \(15\) seeds.
        ${}^{\star}$ indicates statistically significant differences (one-sided independent \(t\)-test) w.r.t.\ the baseline \(\hiddenstate\) \robertabase \, with $p < 0.05$.
    }
\end{table}

\begin{table}[t]
    \begin{center}
        \footnotesize
        \renewcommand{\arraystretch}{1.2}
\begin{tabular}{@{}llll@{}}
    \toprule
      \multicolumn{1}{c}{Input features} 
      & \multicolumn{1}{c}{Precision \(\uparrow\)}
      & \multicolumn{1}{c}{Recall \(\uparrow\)}
      & \multicolumn{1}{c}{F1 \(\uparrow\)}
    \\ 
    \midrule 
    \makecell{\(\hiddenstate\) \robertabase}
       & {\( 47.97 \)} 
       & {\( 56.07 \)} 
       & {\( 51.67 \)} 
    \\[0.3em] 
    \makecell{\(\hiddenstate \; \robertabase\) \\ \(\oplus \text{static } \persistentimage{0}\)}
       & {\( 47.92 \)} 
       & {\( \textbf{56.85} \)} 
       & {\( 51.94 \)} 
    \\[0.7em] 
    \makecell{\(\hiddenstate \; \robertabase\) \\ \(\oplus \text{contextual } \persistentimage{0} \)}
       & {\( \textbf{48.92}_{\star \text{stat}}^{\star \text{LM}} \)} 
       & {\( 56.23 \)} 
       & {\( \textbf{52.24} \)} 
    \\ \bottomrule 
\end{tabular}
    \end{center}
    \caption{
        \label{tab:term_extraction_phrase_level_baseline_lm_hidden_vs_others_cross_validation}
        Cross-validated phrasal-level performance on SGD for term extractors trained on MultiWOZ2.1 training split without selected domain in
        \(
             \left[ 
                \text{attraction};
                \text{hotel}; 
                \text{restaurant};
                \text{taxi};
                \text{train} 
            \right]
        \) averaged over \(10\) seeds for each of the five folds.
        ${}^{\star}$ indicates statistically significant difference with $p < 0.05$, w.r.t.\ the baseline \(\hiddenstate\) \robertabase \, (\(\text{LM}\)) and augmentation with static persistence images (\(\text{stat}\)).
    }
\end{table}

\begin{table*}[ht]
    \begin{center}
        \scriptsize
        \begin{tabular}{llll}
            \toprule
            \makecell{Ground Truth Terms}
            & \makecell{\(\hiddenstate \; \robertabase\) \\ (our baseline)}
            & \makecell{\(\hiddenstate \; \robertabase\) \\ \(\oplus \text{static } \persistentimage{0}\)}
            & \makecell{\(\hiddenstate \; \robertabase\) \\ \(\oplus \text{contextual } \persistentimage{0} \)}
            \\ 
            \midrule
            \makecell{
                {``cafe jolie''} 
            } 
            & \makecell{
                {\truepositive ``cafe jolie''} 
            }
            & \makecell{
                {\truepositive ``cafe jolie''} 
            }
            & \makecell{
                {\truepositive ``cafe jolie''} 
            } 
            \\ 
            \midrule
            \makecell{``angelina jolie''}  
            & \makecell{ 
                {--} 
            }
            & \makecell{ 
                {\truepositive ``angelina jolie''} 
            }
            & \makecell{ 
                {\truepositive ``angelina jolie''} 
            }
            \\ 
            \midrule
            \makecell{ 
                {``yellow chilli''} \\
                {``the yellow chilli by chef sanjeev kapoor''}  
            }
            & \makecell{ 
                {\truepositive ``yellow chilli''} \\
                {\falsepositive ``the yellow chilli''} \\
                {\falsepositive ``the yellow chilli by''} \\
                {\falsepositive ``sanjeev kapoor''}
            }
            & \makecell{ 
                {\truepositive ``yellow chilli''} \\
                {\falsepositive ``the yellow chilli''} \\
                {\falsepositive ``the yellow chilli by chef''}  \\
                {\falsepositive ``sanjeev kapoor''}
            }
            & \makecell{ 
                {\truepositive ``yellow chilli''} \\
                {\falsepositive ``the yellow chilli''} \\
                {\truepositive ``the yellow chilli by chef sanjeev kapoor''} 
            }
            \\ 
            \midrule
            \makecell{ 
                {``water seed''} \\
                {``water seed concert''}
            }
            & \makecell{ 
                {\truepositive ``water seed''} \\
                {\falsepositive ``the water seed''} \\
                {\falsepositive ``water seed event''}
            }
            & \makecell{ 
                {\truepositive ``water seed''} 
            }
            & \makecell{ 
                {\truepositive ``water seed''} \\
                {\truepositive ``water seed concert''}
            }
            \\ 
            \midrule
            \makecell{ 
                {``be alright''} 
            }
            & \makecell{ 
                {--} 
            }
            & \makecell{ 
                {\falsepositive ``alright''} \\
                {\truepositive ``be alright''}
            }
            & \makecell{ 
                {\falsepositive ``alright economy''} 
            }
            \\
            \bottomrule
        \end{tabular}
    \end{center}
    \caption{
        \label{tab:term_extraction_qualitative_analysis}
        Representative examples of predictions where the baseline model fails to retrieve the correct term, while a local topology feature augmented model succeeds.
        We indicate {\truepositive true positives} and {\falsepositive false positives} by color.
    }
\end{table*}

\Cref{tab:term_extraction_phrase_level_baseline_lm_hidden_vs_others_full_data} lists the term extraction performance for the pure language model baseline, our proposed method of augmenting with contextual topological features, and the alternative approach by augmenting with the static topological features from~\citet{vukovic-etal-2022-dialogue}. 
The main objective of said work was the maximization of recall, and to that end they proposed separate language model and topological feature taggers, with a subsequent union of predictions.
In contrast, we show that our unified model augmented with contextual topological features can increase precision, recall, and F1 over the language model baseline.

\Cref{tab:term_extraction_phrase_level_baseline_lm_hidden_vs_others_cross_validation} presents averaged results for the models trained on a reduced dataset constructed by omitting a given domain in the MultiWOZ2.1 training set.
Here, on average, the augmentation with the contextual persistence images is again better than the language model vector baseline.

\paragraph{Qualitative Analysis}

To obtain explicit examples, we select a model checkpoint for each feature type after \numprint{1100} global steps, and inspect the differences between predicted normalized phrase sets.
In \Cref{tab:term_extraction_qualitative_analysis} we see examples where the topologically augmented model succeeds in finding complete multi-word terms, whereas the baseline model cuts off before the end of a term or misses proper names that should follow a preposition.
Such information is highly dependent on the context of a term, and the contextual topological model is able to find long terms more consistently.
All models identify the restaurant name ``Cafe Jolie'' correctly, but only the topological models recognize the actress ``Angelina Jolie''.
Similarly, the song title ``Be alright'' containing the frequent word ``alright'' is not recognized by the language model alone, but can be detected by a topological model.

\subsection{Relation to the Manifold Hypothesis}

At first glance, our results may appear to be at odds with the \emph{manifold hypothesis}, a common assumption underlying many representation learning paradigms.
While this hypothesis has been questioned for \emph{static} word embeddings \citep{jakubowski-etal-2020-topology}, it remains uncontested for contextual embeddings.
It posits that the latent space of a trained machine learning model is clustered around a disjoint union of lower-dimensional manifolds \citep{bengio2013representationlearning,brown2023verifying}.
This implies that, from a purely topological perspective, the local structure of the latent space is constant, at least along connected components -- every point should have a neighborhood topologically identical to an open ball in some Euclidean space.
How then is it possible that we can extract meaningful information from variations of the local topology? 

There are at least two answers to this.  First, all our measures depend on the way data is \emph{sampled}.  There is no reason to assume that the embeddings drawn from a given corpus provide a uniform sample of the latent space.  On the contrary, the distribution of these samples will depend heavily on the corpus.  And within a given corpus, we might expect the neighborhoods of latent vectors of content words to be ``more spread-out'' and of higher dimension than those of non-content words, since there are more plausible possibilities for filling in content words in a text than for non-content words.
Second, our measures are based on persistent homology, which is known to detect not only topological properties but also differentiable structure such as curvature \citep{bubenik2020persistenthomologycurvature}.
Thus, even on a uniformly sampled manifold, these measures are expected to vary.

\subsection{Computational Complexity}

In this section, we address the computational overhead coming from our proposed method of augmenting a sequence tagger with contextual topological information of a given corpus.
The one-off computational costs for the datastore, in our study derived from the MultiWOZ2.1 training dataset, and the query datasets (MultiWOZ2.1 training \& validation dataset, and SGD dataset) involve a single embedding model forward pass for each input sequence.

For each query dataset relative to the datastore, assuming a constant and small neighborhood size \(n\), the asymptotic complexity of the neighborhood search depends on the tokenized cardinality of the query dataset \(|Q|\), the tokenized cardinality of the datastore \(|C|\), and the embedding dimension \(d\). 
The runtime complexity using the exact search implementation from \citep{johnson2019billion} is \(\mathcal{O}(|Q| |C| d)\), and the storage complexity for neighborhood indices is \(\mathcal{O}(|Q| n)\).

The persistent homology computation in dimension \(0\) for each query vector depends on the neighborhood size \(n\) as well. 
For degree \(0\), the number of simplices in the Vietoris-Rips complex can be upper-bounded by \(n^2\). 
Thus, the persistence diagram for each neighborhood can be computed in \(\mathcal{O}(n^{2 \omega})\), where \(\omega < 2.4\) is the matrix multiplication exponent \cite{MR2919613}. 
There are at most \(n\) generators in the \(0\)-dimensional persistence diagrams, so the computation of the Wasserstein norms can be achieved in \(\mathcal{O}(n^3)\) \cite{NEURIPS2018_b58f7d18}. 
Empirically, the computation of the persistence images is observed to be very quick compared to the computation of the persistence diagrams.

Once computed and cached, these topological features can be reused for different training objectives on the given query dataset. 
The only overhead in transitioning from the baseline tagger (approximately 35.65 million parameters) to the tagger with input \(\hiddenstate \, \robertabase \oplus \text{contextual } \persistentimage{0}\) involves a few additional parameters (roughly \numprint{60000}) for the encoding module of the 100-dimensional contextual persistence image. 
Consequently, once the topological features have been cached, the training and inference of the topologically augmented BIO-tagger are only negligibly slower than the baseline BIO-tagger.
\Cref{sec:appendix_computational_complexity_and_implementation_details} discusses the software packages used in the implementation and how we handle caching of the precomputed neighborhoods and resulting contextual topological features.

\section{Conclusion}

In this work, we introduce a topological deep learning approach to enrich feature learning-based sequence tagging methods with contextual topological data.
Our methods do not depend on access to the underlying feature creation method, nor on external knowledge bases.
Once these local topological descriptors are computed, they offer the potential for reuse across different tasks, thereby mitigating the initial computational investment.
One limitation lies in our method still requiring labels on the seed dataset.
Though our results in the case study hint at a correlation between dialogue terms and higher Wasserstein norms, we have yet to establish a clear-cut purely feature based criterion for distinguishing terms from non-terms in dialogue data.

Looking ahead, we conjecture that the utility of our approach extends beyond the term extraction task investigated in this study. 
Given its generic design and challenge, it is plausible that it is applicable to other language models and modalities.
Although our empirical evaluations have been confined to masked language models, the difficulty of the term extraction task provides optimism that our method could be advantageous in other scenarios where understanding the relation between individual data points and a datastore is critical.

\paragraph{Reproducibility Statement}

The MultiWOZ2.1 and SGD datasets are publicly available through the ConvLab-3 unified data format \cite{zhu-etal-2023-convlab}, and we release our preprocessing, local topological feature computation and tagging model training code.\footnote{\url{https://gitlab.cs.uni-duesseldorf.de/general/dsml/tda4contextualembeddings-public}}

\section{Limitations}

The experiments have been confined to a small masked language model (RoBERTa base model).
Our proposed method can be applied to embedding spaces derived from causal LLMs \cite{behnamghader2024llm2vec}, but current state-of-the-art models typically produce latent spaces with significantly larger embedding dimension.
This has great influence on the computational complexity required to generate the contextual topological features.
While our BIO-tagger can be trained on a single V100 GPU with 16 GB of VRAM in 2 hours, one should note that for efficiently precomputing the nearest neighbors in the contextual topological feature extraction, the embedded datastore needs to fit into the graphic card memory.
This one-off neighborhood computation is not an issue for the MultiWOZ2.1 training set datastore, but might limit applications to larger corpus sizes.
One possible remedy could be given by applying embedding space dimension reduction techniques such as \cite{kusupati2022matryoshka} to the datastore before computing our topological features.

\subsubsection*{Acknowledgments}

BMR and RV are supported by funds from the European Research Council (ERC) provided under the Horizon 2020 research and innovation programme (Grant agreement No.\ STG2018 804636) as part of the DYMO project.
CVN and HL are supported by the Ministry of Culture and Science of North Rhine-Westphalia within the framework of the Lamarr Fellow Network.
MH and SF are supported by funding provided by the Alexander von Humboldt Foundation in the framework of the Sofja Kovalevskaja Award endowed by the Federal Ministry of Education and Research.
Computational infrastructure and support were provided by Google Cloud.
We want to thank the anonymous reviewers whose comments improved the quality of our paper.


\bibliography{bibliography}

\appendix
\section{Appendix}
\label{sec:appendix}

\subsection{Implementation Details}
\label{sec:appendix_computational_complexity_and_implementation_details}

In our term extraction applications, the ambient vector datastore comprises a collection of vectors with cardinality in the millions, making the computation of neighborhoods a major computational bottleneck.
To alleviate this issue, we employ the Facebook AI Similarity Search (\Faiss) module \citep{johnson2019billion} to precompute neighborhood indices using GPU acceleration.
These indices can be reused in subsequent computations of our local measurements at varying scales.
We obtain the neighborhood indices and distances for \numprint{1024} neighbors using the \texttt{faiss.IndexFlatL2} build from the MultiWOZ2.1 training datastore.
This neighborhood cache allows extraction of the codensity measurements and the vectors required to subsequently compute persistence images for neighborhood size \(n=128\).
Loading the \numprint{2739744} many 768-dimensional vectors from the \robertabase \, MultiWOZ2.1 training datastore into the \Faiss \, index requires approximately 8 GB of GPU memory.

The \Faiss \, library does not currently offer GPU support for range-based nearest neighbor search.
This makes it infeasible to compute range-based neighborhoods at the scale of our dataset for the methods described in \citet{vonRohrscheidt23a}.
This limitation is especially critical because our BIO-tagger requires topological features for each input token in each context.

Another computational challenge lies in the local persistent homology computations, which become a bottleneck when the goal is training a BIO-tagger based on the resulting features.
To address this, we precompute and store the topological features and their vectorizations, including both persistence images and Wasserstein norms.
We use the Ripser library \citep{Bauer2021Ripser} for computing the persistence modules for \(H_{0}\) and \(H_{1}\) with \(\mathbb{F}_{2}\)-coefficients w.r.t. cosine distance from the precomputed neighborhoods, and GUDHI \citep{gudhi:urm,gudhi:PersistenceRepresentations} for the vectorization and persistence representation.
The Wasserstein norms, i.e., the order-$1$ Wasserstein distances between the neighborhood persistence diagrams and the empty diagram with Euclidean ground metric, are computed separately for the $H_{0}$ and $H_{1}$ persistence diagrams using the GUDHI library.

For the persistence image vectorization of the \(H_{0}\)-persistence module, we decide on a bandwidth of \(0.01\), image range on the \(y\)-axis of \([0.0, 1.0]\), resolution of \(1 \times 100\) and weight each persistence homology generator by its \(y\)-value.
We restrict our computations to 0-dimensional and 1-dimensional persistent homology.
This is not only due to Ripser's optimizations, which result in a faster runtime, but also to circumvent the potential exponential increase in the number of simplices in the filtered complex when considering higher dimensional Vietoris-Rips complexes on a point cloud.

\end{document}